\documentclass{article}

\usepackage{arxiv}

\usepackage[utf8]{inputenc} 
\usepackage[T1]{fontenc}    
\usepackage{hyperref}       
\usepackage{url}            
\usepackage{booktabs}       
\usepackage{amsfonts}       
\usepackage{amsmath}
\usepackage{amssymb}
\usepackage{nicefrac}       
\usepackage{microtype}      
\usepackage{lipsum}
\usepackage{graphicx}
\usepackage{xltabular, multirow}
\graphicspath{ {./images/} }
\usepackage{float}
\usepackage{svg}

\usepackage{subcaption}
\usepackage[round]{natbib}   
\usepackage{color} 
\usepackage{todonotes} 
\usepackage{mathtools}
\usepackage{multirow}
\usepackage{graphicx}
\usepackage{gensymb}

\title{Neural Associative Skill Memories for safer robotics and modelling human sensorimotor repertoires}
\author{%
Pranav Mahajan$^{1}$\thanks{Correspondence: pranav.mahajan@ndcn.ox.ac.uk} \quad Mufeng Tang$^1$ \quad T. Ed Li$^{1,2}$  \quad Ioannis Havoutis$^1$ \quad Ben Seymour$^1$ \\
$^1$University of Oxford \quad $^2$Yale University\\
}

\begin{document}
\maketitle


\begin{abstract}


Modern robots face challenges shared by humans, where machines must learn multiple sensorimotor skills and express them adaptively. Equipping robots with a human-like memory of how it feels to do multiple stereotypical movements can make robots more aware of normal operational states and help develop self-preserving safer robots. Associative Skill Memories (ASMs) aim to address this by linking movement primitives to sensory feedback, but existing implementations rely on hard-coded libraries of individual skills. A key unresolved problem is how a single neural network can learn a repertoire of skills while enabling fault detection and context-aware execution. Here we introduce Neural Associative Skill Memories (ASMs), a framework that utilises self-supervised predictive coding for temporal prediction to unify skill learning and expression, using biologically plausible learning rules. Unlike traditional ASMs which require explicit skill selection, Neural ASMs implicitly recognize and express skills through contextual inference, enabling fault detection across learned behaviours without an explicit skill selection mechanism. Compared to recurrent neural networks trained via backpropagation through time, our model achieves comparable qualitative performance in skill memory expression while using local learning rules and predicts a biologically relevant speed-accuracy trade-off during skill memory expression. This work advances the field of neurorobotics by demonstrating how predictive coding principles can model adaptive robot control and human motor preparation. By unifying fault detection, reactive control, skill memorisation and expression into a single energy-based architecture, Neural ASMs contribute to safer robotics and provide a computational lens to study biological sensorimotor learning.


\end{abstract}


\section{Introduction}

Animals spend their lives learning, storing, and refining sensorimotor skills, which are essential for body estimation and reactive control. A key aspect of effective behaviour is fault detection and fault-tolerant control, which requires an understanding of what normal actions feel like, in order to recognise abnormalities. To achieve this, roboticists have proposed the concept of Associative Skill Memories (ASMs) \citep{pastor2012towards, pastor2013dynamic}. ASMs link attractor models for motor behaviour, such as dynamic movement primitives (DMPs) \citep{ijspeert2013dynamical}, with stereotypical sensory events, creating a memory of typical sensory feedback during movements. DMPs represent and generate movements by encoding the underlying patterns of motion. ASMs build upon DMPs by adding sensory associations, enabling the prediction of sensory events that occur during a movement. This association enhances robustness to perturbations through reactive control, facilitates failure detection akin to anomaly detection, and enables behaviour switching. ASMs can be learned from demonstrations, where an agent observes and replicates movements, then associates the movements with the corresponding sensory feedback.

However, existing ASMs rely on a hard-coded (dictionary-like) library of individual movement primitives, where each primitive has a separate set of weights trained to model the dynamics of each movement from a dataset of demonstrations. Since ASMs memorise each skill separately, the robot requires explicit instruction to choose a specific skill to execute from the library or must manually compare the sensory observation profile of the current skill against those in its library to identify the closest match. Consequently, ASMs can only detect faults or abnormalities with respect to the current skill under execution, correcting them reactively.

In contrast, humans and other animals do not store skills in a hard-coded library. Instead, they utilise neural networks with local learning rules, allowing them to detect abnormalities and correct them reactively and seamlessly. This raises the unresolved question of how a biological or artificial neural network could implement an associative skill memory that captures a repertoire of multiple skills. Enabling a single network to learn multiple skills also presents two key challenges: (1) how to detect abnormalities with respect to all learned movements, without explicit instructions indicating which skill is being executed, and (2) how to decide which skill memory to express or recall during execution.

Predictive coding \citep{rao1999predictive, friston2005theory, clark2013whatever}, an influential conceptual framework in neuroscience, is known to account for several findings of predictive processing in the brain \citep{friston2009predictive, friston2018does}. Predictive coding posits that the cortex performs inference and learning on a hierarchical probabilistic generative model, which learns to predict incoming sensory signals in a self-supervised manner. Recently, it has also been proposed as a biologically plausible alternative approximating backpropagation \citep{whittington2017approximation, whittington2019theories}, in some instances outperforms backpropagation \citep{song2024inferring} and has been applied to the problem of associative memories \citep{salvatori2021, tang2024sequential}. However, its application to the problem of associative skill memories or learning from a demonstration setting has not been explored. Recently, predictive coding-based embodied approaches to a related problem, body state perception \citep{lanillos2018adaptive}, have been shown to perform multisensory integration from noisy information to filter internal state. However, their learned forward model did not account for robot dynamics, which is crucial in differentiating between different movements. Thus, prior work on predictive coding for body perception is limited to online filtering, while we aim to solve the problem of ASMs i.e. learning a complete sensorimotor repertoire (i.e., multiple skill memories at once) and using it for fault detection and fault-tolerant control.

Here, in our neurorobotics simulation-based study, we reduce sensorimotor memory to sequential memory from demonstrations, eschewing optimal planning of motor actions that lead to memorised sequences. Using embodied generative models to capture the robot's dynamics, our temporal predictive coding network learns in a self-supervised manner with biologically plausible, local learning rules. We preview its structure and working before demonstrating its applications in safe self-preserving robotics — fault detection, reactive control, and human motor control - contextual inference in skill expression \citep{sheahan2016motor}. 

\section{Results}

\subsection{Associative Skill Memories (ASMs) and Neural ASMs}

In animals, associative memory retrieves detailed past patterns from partial cues. Neural ASMs, inspired by this principle, retrieve sensorimotor skills from incomplete queries but differ from traditional ASMs originally defined by \citet{pastor2012towards, pastor2013dynamic}. ASMs augment dynamic movement primitives (DMPs) with sensory associations, requiring multiple handcrafted modules that interact (Fig. \ref{fig1}A). Specifically, an explicit skill library keeps track of DMPs, a separate associative memory system maintains signal statistics per movement, and a sensor prediction module compares current sensor signals to the average of previously experienced signals to retrieve the closest-matching DMP from the (dictionary-like) skill library.

Neural ASMs replace all these hand-crafted modules with a single predictive coding network for temporal prediction (tPC) \citep{millidge2024, tang2024sequential} (see Fig. \ref{fig1}B), optimising a unified energy function (equation \ref{eqn:energy_f} or \ref{eqn:energy_f2}) and learning through local Hebbian rules (equations \ref{eqn:learn_WH}, \ref{eqn:learn_WF}). Designed to emulate brain-like learning, Neural ASMs use a tPC network with a Hidden Markov Model (HMM) structure (Fig. \ref{fig1}C), where hidden states $z$ and observations $x$ interact through weights $\mathbf{W}_F$ (predicting observations from hidden states) and $\mathbf{W}_H$ (predicting transitions between hidden states). The inputs to Neural ASMs are sensorimotor observation sequences from demonstrations and the network learns from these demonstrations.

\subsection*{System overview}

The temporal predictive coding network (tPC) operates in discrete time steps, $(\mu = 0, 1,..., T-1)$, abstracting continuous passage of time. Each skill is a sequence of sensorimotor observations, and the network's aim is to predict the next sensorimotor observation at each step. The inputs at time step $\mu$ include sensory observations $x_s$ (here, sensed gripper force/torque, visual cues) and proprioceptive inputs $x_m$ (here, joint angles, end-effector position). The robot can be controlled in configuration space (joint angles) or task space (end-effector position), with all simulations using position control.

\begin{figure}[H]
\centering
\includegraphics[width=0.9\textwidth, keepaspectratio]{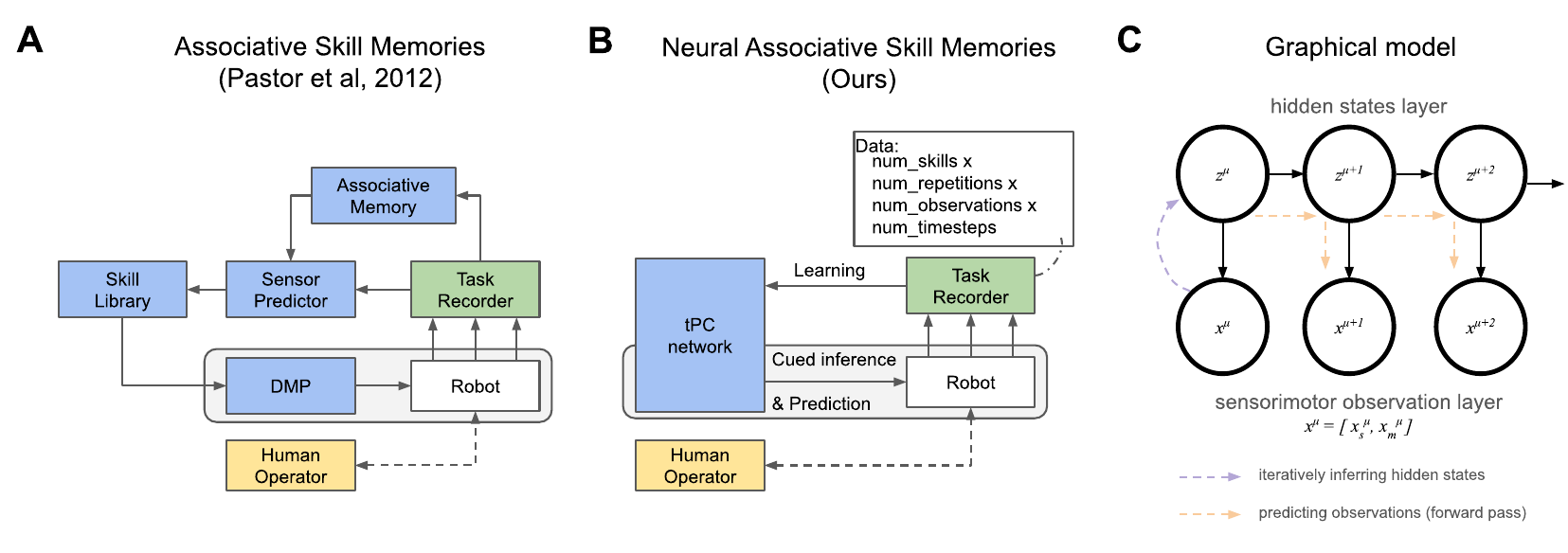}
\caption{(A \& B) Comparison of our Neural ASM model using temporal predictive coding (tPC) network with ASMs \citep{pastor2012towards} which uses Dynamic Movement Primitives (DMPs). Our model replaces the hard-coded library of movement primitives in ASMs with a single tPC network responsible for learning multiple skill memories. The input to neural ASMs is multiple repetitions or demonstrations of multiple skills, where a skill demonstration is a time series of sensorimotor observations (C) Graphical model tPC is that of a Hidden Markov Model (HMM). $x$ represents sensorimotor observations, whereas $z$ are the hidden states which capture the dynamics and the inferred context. Here, we show offline recall or prediction of a sequence of future sensorimotor observations using only the first sensorimotor observation $x^\mu$ as input during cued inference. It is also possible to use observations from multiple time steps during cued inference, which is not shown here. In an HMM, the prediction step is equivalent to a forward pass (this may not always be the case and should be performed by iterative energy minimisation in such scenarios).} \label{fig1}
\end{figure}

The tPC network operates in two phases: (1) memorization and (2) recall. During memorization, the network learns from sensorimotor sequence demonstrations recorded during task performance (Fig. \ref{fig1}B). Each skill data includes multiple, imperfect repetitions, supporting generalization across variations. Learning involves iteratively inferring the hidden state \(z^\mu\) by minimizing the energy function for a set number of iterations (equation \ref{eqn:iterative_inf}) and then updating weights ($\mathbf{W}_H$) and ($\mathbf{W}_F$) (equations \ref{eqn:learn_WH}, \ref{eqn:learn_WF}). This process, termed "prospective configuration" \citep{song2024inferring}, has been shown to have benefits over backpropagation in biologically relevant tasks. Weight updates adapt the energy landscape, creating attractor states for memorized sequences \citep{salvatori2021}, which is crucial in recognising memorised skills.

In the recall phase (memory expression), the network retrieves sensorimotor sequences from partial cues, predicting future observations (equation \ref{eqn:iterative_pred}). This recall process further involves two processes: (1) cued inference - where the hidden state is inferred by integrating multiple sensorimotor observations (Fig. \ref{fig1}C, purple arrow); and (2) prediction -  generating sensorimotor predictions for subsequent time steps (Fig. \ref{fig1}C, orange arrows). 
The combination of these processes results in two kinds of recalls - offline and online recall. In offline recall, cued inference of the hidden state is performed from one or more time steps of sensorimotor observation to essentially set the initial state of the neural dynamics, which then allows to predict the future observations. Similar processes have been hypothesised the role motor preparatory activity in setting the initial state of a dynamical system rather than explicitly represent movement parameters \citep{churchland2010cortical, churchland2006neural, churchland2006preparatory, cisek2006preparing, fetz1992movement}. Prior work demonstrates also the use of such characteristic of initial sensitivity in recurrent neural networks to generate multiple sensorimotor sequences \citep{nishimoto2004learning, nishimoto2008learning}, via association between initial states and corresponding sequences. In offline recall, cued inference of hidden state is not invoked during the prediction process thus the trajectory or motor plan remains fixed post-cued inference.  Alternatively, in online recall, the model continuously performs cued inference of hidden state making sensorimotor prediction for the next time step supports adaptive skill re-recognition and re-planning to changing inputs. Offline recall is a model of (1) ballistic movements on a shorter timescale where it is expensive to run iterative inference, or (2) a sequence of movements guided from working memory \citep{mizes2023dissociating,mizes2024role} on a longer timescale and (3) mental simulation, where the robot autonomously generates sensorimotor trajectories without producing actual movements \citep{yamashita2008emergence, jeannerod1994representing, tani1996model}. Whereas online recall is a model of usual movements which are guided by and require sensory inputs at every time step \citep{mizes2023dissociating,mizes2024role}. Animals when awake, may employ further techniques to further optimise the efficiency but we do not cover these techniques. Examples include an adaptive interpolation of the two recalls such as scaling the number of inference iterations proportional to the surprise or alternatively, utilising an amortised inference \citep{dayan1995helmholtz, hinton1995helmholtz, gershman2014amortized}. We provide further details on the learning rules and both recall approaches in the Methods section, along with a neural network illustration, but all simulations in this paper utilise offline recall.

To achieve desired states in configuration or end-effector space, the system relies on a hardcoded low-level controller, which is often a problem separately studied under learning inverse models \citep{wolpert1995internal, wolpert1998multiple}. Neural ASMs do not prescribe a specific low-level control approach; instead, they function as a high-level dynamical policy, guiding a flexible control system (cf. \citet{schaal2007dynamics}) (Appendix 1, Fig. \ref{schaal_appendix}). Regardless of recall mode (offline or online), the low-level controller operates online, allowing reactive adjustments toward a desired state even when set through offline recall. We will next see how Neural ASMs can be used for fault detection and reactive correction.

\subsection{Memorised skills are useful in fault detection and simple reactive correction}

\begin{figure}[H]
\centering
\includegraphics[width=\textwidth, keepaspectratio]{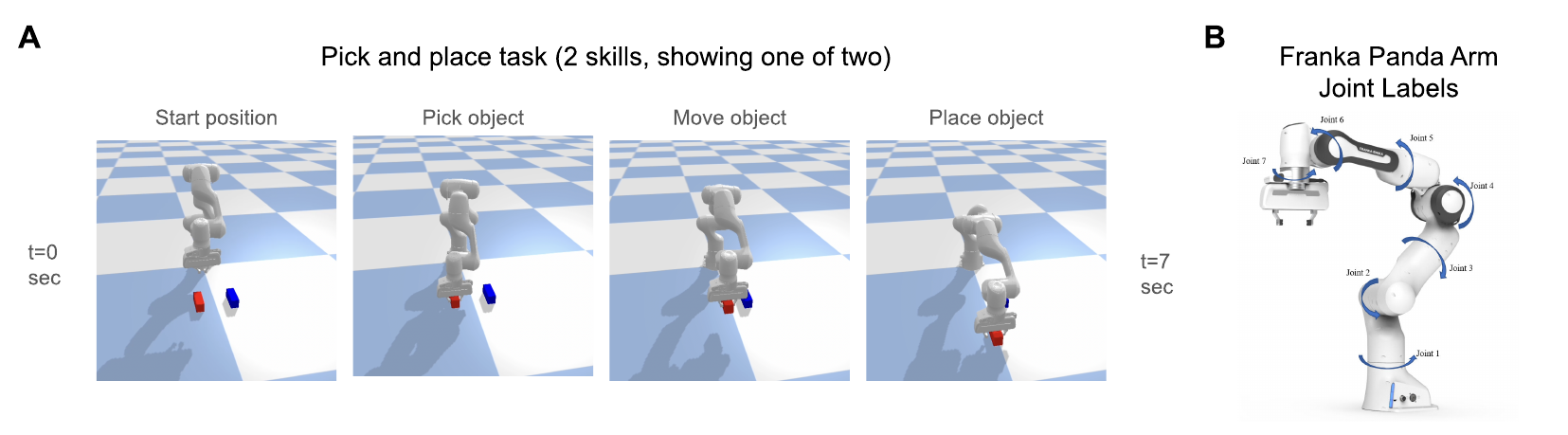}
\caption{(A) Demonstration of Neural ASMs learning two pick and place skills in simulation. The sensorimotor sequences in the dataset used for learning from demonstrations are generated using predefined end-effector goals and use inverse kinematics to get joint angles. This is intended to be a proxy for teleoperation in simulation. (B) A schematic of Franka Panda arm joint labels (adapted from \citet{rogel2022robogroove})} \label{fig2}
\end{figure}

Having introduced Neural ASMs as a viable alternative to ASMs, we now demonstrate their core functionalities: fault detection and reactive fault correction in a simple simulation. To provide the simplest demonstration of learning multiple skills, we trained Neural ASMs to memorize two pick-and-place skills (Fig. \ref{fig2}A) from 10 demonstrations of each skill. Further details on training and task simulation are in the Methods section. Each skill had distinct starting observations, allowing the network, once trained, to infer the skill based on initial observations alone and then predict the sequence of sensorimotor observations using offline recall. The robot executed the skill using configuration-space control. This process of skill retrieval relies on implicit skill recognition during cued inference, unlike ASMs \citep{pastor2012towards, pastor2013dynamic}, which require the explicit selection of a dynamic movement primitive (DMP).

\begin{figure}[H]
\centering
\includegraphics[width=\textwidth, keepaspectratio]{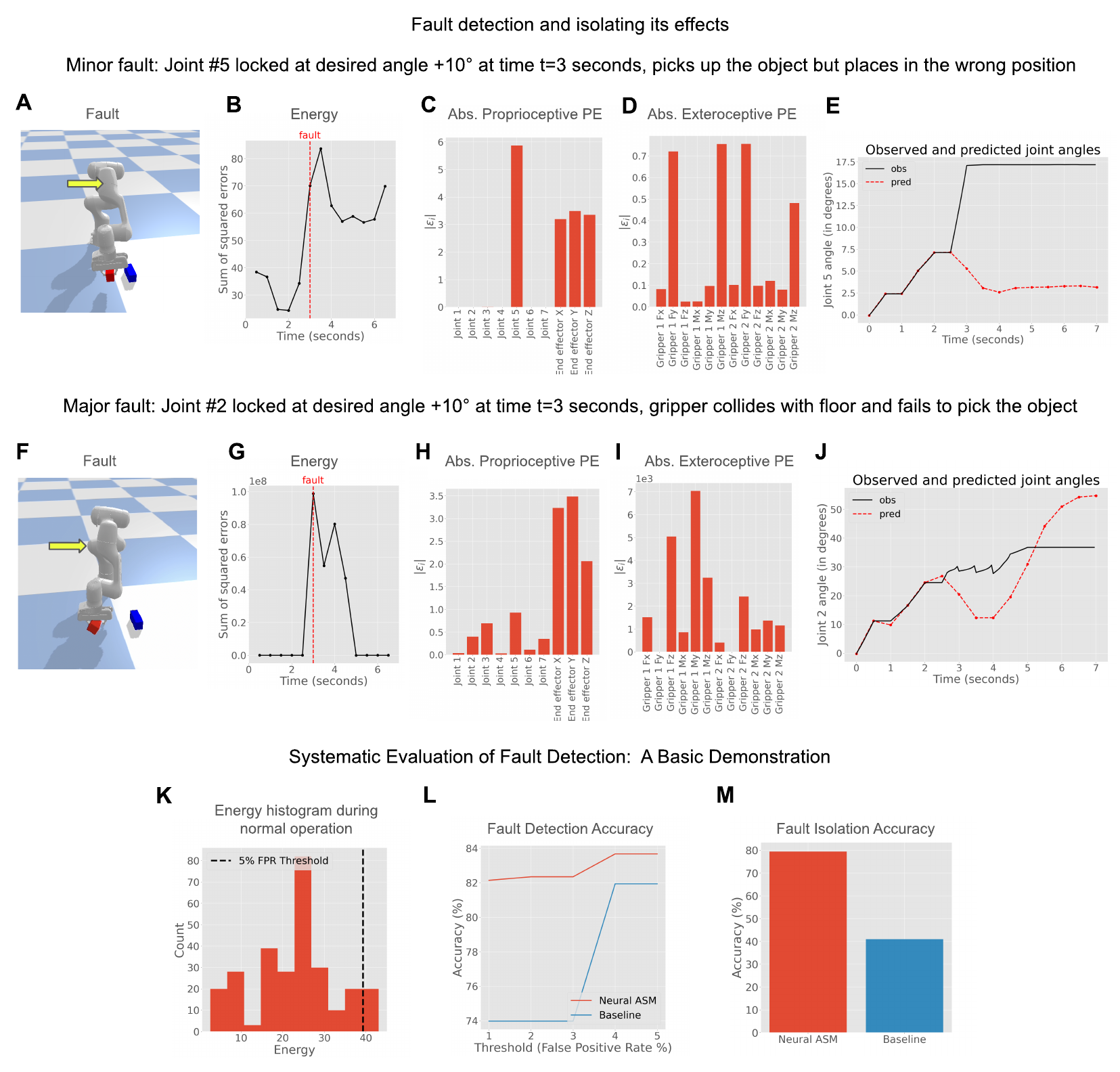}
\caption{(A-E) Minor fault example with fault detection using energies and correct fault isolation using absolute prediction errors along with the joint angle time series. (F-J) Major fault example with fault detection using energies and incorrect fault isolation using absolute prediction errors along with the joint angle time series. (K-M) A basic demonstration of systematic evaluation of fault detection.} \label{fig_fd}
\end{figure}

We next demonstrate fault detection by measuring the energy of the network (equation \ref{eqn:energy_f2}). Similar methods have been employed in predictive coding networks (without temporal prediction) to detect novelty in stationary images \citep{li2024predictive}. In our robot experiment, once the cued inference stage is complete during the offline recall process, this reduces to simply the sum of squared errors in the observation layer. An example of a common fault with robotic arms is an arm getting locked in a position, e.g. a robot malfunction or someone pushing the robotic arm. We, therefore, first demonstrate fault detection and isolation in two examples of joint-locking and then provide a systematic evaluation of fault detection. A diagram of Franka Panda arm joint labels is provided in Fig. \ref{fig2}B to help follow the examples. In our first example, the fault is simulated as follows: the joint 5 overshoots the desired goal by 10$\degree$ while attempting to pick up the object ($t=3$ seconds) and gets stuck in that configuration for the remaining duration (Fig. \ref{fig_fd}A). The fault is detected using abnormal (out-of-distribution) energies at $t=3$ seconds, demonstrating fault detection (Fig. \ref{fig_fd}B). Since the effect of the fault is limited (the object gets placed in the wrong location), we can reliably tease out the effect of the fault (which here is also the cause), i.e joint 5 has the highest absolute proprioceptive prediction error (Fig. \ref{fig_fd}C). We also report absolute exteroceptive prediction errors corresponding to forces and torques on the two grippers (Fig. \ref{fig_fd}D) and plot the joint 5 angle over time for further clarity (Fig. \ref{fig_fd}E). In our second example, we simulate a fault where joint 2 overshoots the desired goal by 10$\degree$ and gets stuck while attempting to pick up the object, which results in the arm colliding with the floor (Fig. \ref{fig_fd}F). The fault is detected using abnormal (out-of-distribution) energies at $t=3$ seconds, demonstrating fault detection  (Fig. \ref{fig_fd}G). One can further utilise the individual absolute prediction errors to isolate the effects of the fault, though it might not always help isolate the causes. Here, despite narrowing down to joint angles (because of configuration-space control), we observe higher downstream effects of a fault in joint 2 in joint 5 angle, joint 3 angle, the end-effector co-ordinates and gripper sensory observations due to collision of the gripper with the floor (Fig. \ref{fig_fd}H and I). We also plot the joint 5 angle over time for further clarity (Fig. \ref{fig_fd}J). Thus, we have demonstrated two examples of fault detection and isolation of its effects, in one example, the cause is the same as the effect, whereas in another, it is not. A subtlety about Neural ASMs is that since the skill recognition is entirely implicit, the fault detection only depends on the current predictions, which in turn depends on the inferred context. This is unlike ASMs, which require knowing the explicit movement being performed to compare the observations to the corresponding signal statistics.

We next present a more systematic evaluation of fault detection using Neural ASMs, however, it is a basic demonstration. For simplicity, we use percentile-based thresholding of the energy distribution during normal operation without faults to set the threshold for fault detection. Fig. \ref{fig_fd}K shows the energy distribution during our pick-and-place task over 10 trials of normal operation for each skill along with the 95th percentile threshold as an example, which is equivalent to a 5\% false positive rate (FPR). Next, we vary the threshold from 99th percentile to 95th percentile and thereby vary the false positive rate from 1\% to 5\% and then measure the accuracy of fault detection on simulated faults. The failure to detect faults from a range of simulated faults is the false negative rate (FNR), which is equal to (100 - accuracy)\% in our example. Therefore, the set threshold controls for FPR, and we can measure the accuracy or FNR in fault detection.  Accuracy in fault isolation is measured based on the matches between the joint where the fault was simulated and the joint which had the highest absolute prediction error on the timestep of the fault.

We simulate 980 trials with faults by varying the joint that gets locked (joint 1 to 7), the time of fault (t=1 to t=6) and the degree of joint angle overshoot (-15$\degree$ to +15$\degree$). We further design a simple baseline, which, instead of using prediction errors, uses normalised errors similar to Z-score normalisation for each observation channel, i.e. subtracts the mean value and divides by standard deviation. This is analogous to one possible implementation by traditional ASMs, which store signal statistics such as mean and standard deviation over multiple demonstrations (referred to as associative memory component in their framework by \citep{pastor2012towards}). However, we are mindful that ASMs propose a general recipe for fault detection, isolation and correction rather than prescribing a specific method, therefore, we refer to this as a baseline to compare against, and better ways to use summary statistics over recorded observations may exist. For purposes of comparison, we then also set the fault detection threshold using the sum of squared errors using the baseline method, and fault isolation was similarly performed by identifying the joint with maximum normalised errors. Further details are provided in the Methods section - Robot simulation details. We find that Neural ASMs detect 82-83\% of faults, outperforming the baseline, which detects 74-82\% of faults for a reasonable range of thresholds (1-5\% FPR). We also find that Neural ASMs correctly isolate the faults in 79.5\% of cases, outperforming the baseline accuracy of 41\%. This concludes our basic demonstration of systematic evaluation of fault detection and isolation, which can be extended to real-world robots with more realistic faults in future work.

\begin{figure}[H]
\centering
\includegraphics[width=\textwidth, keepaspectratio]{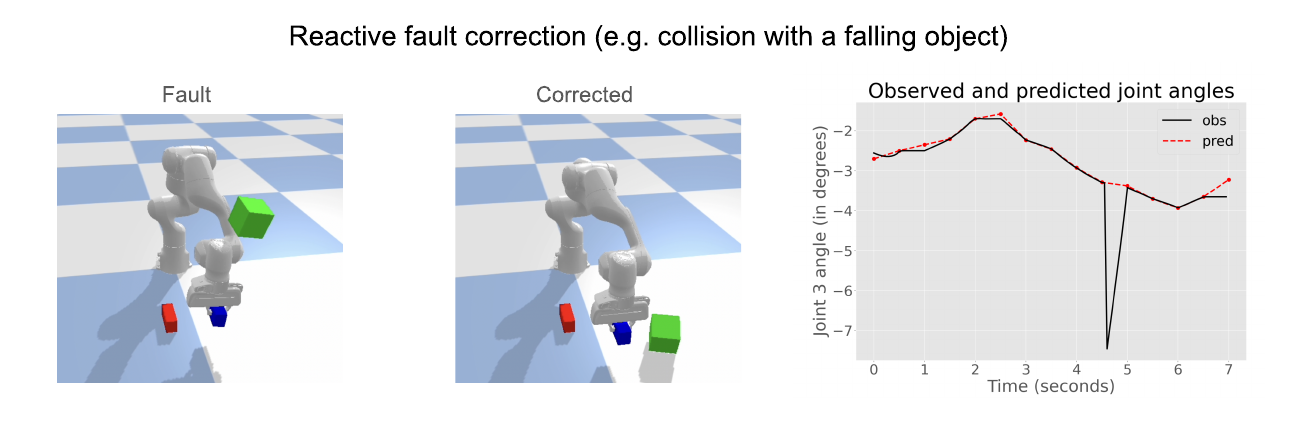}
\caption{Demonstration of a fault resulting from a collision with a falling object. The fault is corrected reactively on the fly by having the low-level controller minimise proprioceptive prediction errors in joint configuration space using the proprioceptive predictions from Neural ASMs.} \label{fig_rc}
\end{figure}

Lastly, in cases where faults can be corrected on-the-fly, Neural ASMs enable reactive correction. In ASMs, this is facilitated by DMPs which themselves provide the reactive movement dynamics in end-effector space. In Neural ASMs, reactive correction is modelled by minimising proprioceptive prediction errors in either end-effector or configuration (joint) space, which is being used for control. We demonstrate reactive fault correction by simulating a fault caused by a falling cube colliding with the robot, leading to temporary prediction errors. By minimizing proprioceptive prediction errors in configuration-space based on predicted trajectory, the low-level controller automatically corrected for this disturbance (Fig. \ref{fig_rc}). In this simulation experiment, almost all faults can be corrected on-the-fly unless the grip strength is too weak and the object slips out of grip due to the collision. A more systematic evaluation will require extending Neural ASMs to real world robots with along with alternative human-like methods for fault-correction e.g. \citet{collins2005efficient}, please see Discussion section for more details.

In summary, this simple setup showcases that the core aspects of ASMs can be implemented using Neural ASMs: (1) fault detection, enabling the robot to halt and seek assistance for unresolvable faults, and (2) reactive fault correction, supporting real-time adjustments for robust, fault-tolerant control.

\subsection{Contextual inference in skill memory separation and expression}

Having demonstrated the utility of Neural ASMs for self-preserving robots, we now demonstrate the role of contextual inference in the separation and expression of skill memories in our model. 
We utilise a robotics setup loosely inspired by \citet{sheahan2016motor}, who showed that motor planning of a follow-through motion, but not simply its execution, separates sensorimotor memories. However, we radically simplify the setup and eschew the optimal control that goes into arriving at the optimal trajectories which compensate for the perturbations applied in their experiment. We rather assume that the appropriate sensorimotor sequences comprising the motor plan for each context are available in the demonstration dataset (for learning from demonstrations) and focus solely on under which conditions our model can or cannot learn and express these memories. We aim to explain and qualitatively simulate certain aspects of human motor behaviour in this robotics task. In doing so, we will also compare Neural ASM with less biologically plausible counterparts (baselines): sequence-to-sequence recurrent neural networks (RNNs) trained using backpropagation through time (BPTT). This simulation highlights the challenges shared by humans and machines that utilise neural network learning multiple skills and would not usually arise if the system stored each skill independently in a library-like manner.  

\begin{figure}[H]
\centering
\includegraphics[width=\textwidth, height=\textheight, keepaspectratio]{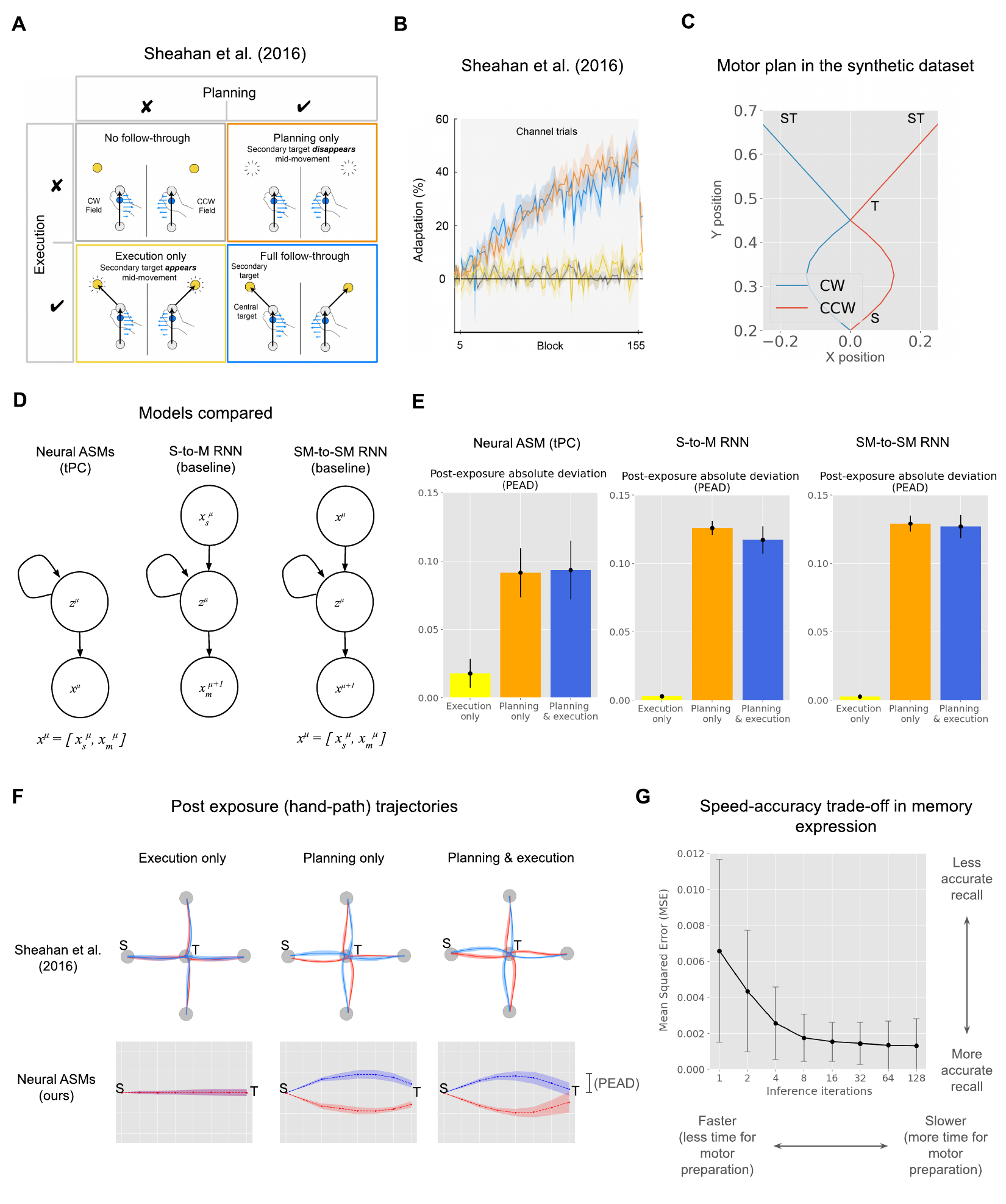}
\caption{(A) Schematic from \citet{sheahan2016motor} describing their experiment which inspires our robotics experiment (B) Adaptation result from \citet{sheahan2016motor}(C) Motor plan (end-effector positions) in the synthetic dataset, S: starting point, T: central target, ST: secondary target. Further details in the Appendix 2, Fig \ref{sheahan_appendix} (D) Simplified representations of our model and baseline RNNs. S-to-M RNN predicts the motor observations at the next discrete time step ($\mu +1$) using the sensory observations at discrete time step ($\mu$) as input. SM-to-SM predicts the sensorimotor observations at the next discrete time step ($\mu +1$) using sensorimotor observations at the current time step ($\mu$) (E) Results qualitatively replicated by our model, on par with baseline models. (F) Qualitative comparison of post-exposure (hand-path) trajectories (G) Speed accuracy trade-off predicted in-memory expression. Here, demonstrated in offline skill recall.} \label{fig3}
\end{figure}

Specifically, the experiment involved reaching a target whilst counteracting the motor perturbations exerted by a manipulandum, either in the clockwise (CW) or counter-clockwise (CCW) direction. The participants in different groups were asked to either plan and execute a follow-through to a secondary target (Planning and execute condition), plan but not execute the follow-through (Plan only condition), execute a follow-through without planning it before movement initiation (Execute only condition) or neither plan nor execute the follow-through. Crucially, the visual cue corresponding to the secondary target was associated with the direction of the force field (Fig. \ref{fig3}A). Participants engaged in motor preparation (for 300 ms) prior to the start of the movement, performing multisensory integration (e.g. using visual cues of secondary target if available prior to the movement). In the Planning only condition and Planning and execution condition, the secondary target appeared prior to the movement, whereas in the execution-only condition, the secondary target appeared during the movement and was not available prior to the movement (Fig. \ref{fig3}A). They found adaptation in test trials (channel trials) and also in post-exposure hand trajectories of participants learned to counteract the perturbation in planning groups but not in the execution group (Fig. \ref{fig3}B). They measure adaptation in channel trials as the slope of the regression of the time course of the force that participants produced into the channel against the ideal force profile that would fully compensate for the field. The different post-exposure hand trajectories for opposing perturbations showed that the participants in planning groups had learned to separate the two skill memories responsible for counteracting the respective perturbative forces. 

Inspired by \citep{sheahan2016motor}, we design a simulated robotics setup with the synthetic data for the three conditions with the same motor plan to compensate for (hypothetical) perturbations while reaching the primary target and then followed through to the respective secondary target (shown using end-effector position coordinates since we use position control Fig. \ref{fig3}C). We do not provide human experiment-like perturbations in the simulation, as the goal of this simulation experiment is not to demonstrate the optimal control necessary to counteract the forces, but rather to show under which conditions the model can separate and express different skill memories. Therefore, we directly concern ourselves with different motor plans available in the form of demonstrations in the dataset. The synthetic dataset for each skill includes end-effector positions and the joint angles calculated using inverse kinematics for planned and observed movements and a one-hot coded visual cue. The synthetic dataset is plotted in Appendix 2 (Fig. \ref{sheahan_appendix}). We do not model the fourth, "no follow-through" condition; further explanation is provided in the discussion section.

We trained Neural ASMs using the tPC network on this synthetic robot data within learning from the demonstration framework and compared them with baseline RNN models (Fig. \ref{fig3}D). The baseline sensory-to-motor (S-to-M) RNN is a discrete-time recurrent network mapping sensory inputs to motor outputs. The baseline sensorimotor-to-sensorimotor (SM-to-SM) is a discrete-time variant of continuous-time sensorimotor RNNs used in \citep{nishimoto2008learning}. Note that, unlike tPC networks, these baseline RNNs are trained using BPTT and do not have the iterative inference phase. The iterative cued inference of the hidden state on the first time step is a model of the neural activity during motor preparation, here hypothesised to set the initial state of the dynamical system \citep{churchland2010cortical, churchland2006neural, churchland2006preparatory}. The Neural ASM model performs offline recall to express skill memories, assuming the movement is a ballistic movement (since the complete trial took less than 1.5 seconds, with at least 300 ms for motor preparation \citep{sheahan2016motor}). Similarly, the baseline models mimic the offline recall, utilising inputs only from the first time step. We train all networks with 6 different seeds separately for each condition, for 1200 learning trials and post-training offline recall is averaged over 24 trajectories. This resembles the 6 participants, 1200 exposure trials and 24 post-exposure trials used in the human experiment.

We find that in the Planning only and Planning and Execution conditions, the model can separate the two skill memories and express them, but not in the Execution only condition (Fig. \ref{fig3}E). This is because, in the Execution only condition, the sensory observations (visual cues) to distinguish the two skill memories are unavailable, and the contextual inference is unable to recall the correct motor plan. We measure the ability to separate memories during skill expression in terms of Post-exposure absolute deviation (PEAD) (Fig. \ref{fig3}F), i.e. how much do the trajectories deviate from the straight line towards the target during test time? We define PEAD as the maximum absolute deviation of the recalled motor plan perpendicular to the line from the starting point (S) to the central target point (T), see Fig. \ref{fig3}F. Since our learning from demonstrations dataset has the optimal trajectories, if the model can separate the skill memories under a condition, then we expect to see the trajectories deviate in opposing directions for different cues for CW or CCW perturbation (denoted by red and blue in Fig. \ref{fig3}F), similar to our synthetic dataset (Fig. \ref{fig3}C). Our setup doesn't simulate the robot learning to counteract perturbations, therefore, we think PEAD is an analogous proxy measurement for the adaptation measured by \citet{sheahan2016motor} (Fig. \ref{fig3}B).  We find that the Neural ASM model, as well as the baseline RNN models can qualitatively replicate the post-exposure adaptation results by \citet{sheahan2016motor} (Fig. \ref{fig3}E). This can also be seen in the recalled motor plan, where we find qualitatively similarity with the post-exposure hand path trajectories in \citet{sheahan2016motor}(Fig. \ref{fig3}F). \citet{sheahan2016motor} visualise their post-exposure hand path trajectories with 4 different starting points on the outside and ending in the central target,  whereas we simplify when plotting our recalled motor plan in Fig. \ref{fig3}F.

Interestingly, our model predicts a speed-accuracy trade-off in such ballistic actions which is not predicted by baseline RNNs (Fig. \ref{fig3}G). Unlike baseline RNNs, our model has a cued iterative inference phase akin to setting the initial condition of the hidden state from multi-sensory integration. Since we hypothesise such iterative inference of hidden state as a potential mechanism of motor preparation, fewer inference iterations mean a shorter or constrained time of motor preparation. We find that very few iterations may be inadequate for complete hidden state inference, leading to less accurate skill recall, which improves with an increase in inference iterations. This additional timescale in our model predicts such speed-accuracy recall during skill recall or expression. A similar speed-accuracy trade-off has been observed in the expression of habitual actions in human experiments by \citet{hardwick2019time}, please refer to the Discussion section for more details.

We have demonstrated that our model can capture essential aspects of the results by \citep{sheahan2016motor} on the role of contextual inference in sensorimotor memory separation and expression in our simplified robotics setup and is qualitatively on par with RNN baselines using non-local learning rules. Further, it predicts a speed-accuracy trade-off during skill memory expression, which is not predicted by baseline RNNs.

\section{Discussion}

This work proposes Neural Associative Skill Memories (ASMs), a framework for learning embodied generative models of sensorimotor repertoires using self-supervised learning and local learning rules. Unlike traditional ASMs, which train separate models for each skill, our approach leverages a single predictive coding network for temporal prediction, allowing multiple skills to be learned, expressed, and updated within the same network. As an energy-based model, Neural ASMs support out-of-distribution detection for fault identification and minimize proprioceptive prediction errors for reactive control. Our findings show that this framework qualitatively captures the role of contextual inference in memory separation and expression in a simplified robotics setup. When it comes to explaining these experimental findings, it demonstrates comparable recurrent neural network (RNN) baselines trained using BPTT \citep{williams1990efficient}, while relying on biologically plausible local learning rules.

\subsection*{Relation to previous neuroscientific experiments, computational models and novel predictions}

Neural Associative Skill Memories (ASMs) relate to neuroscientific frameworks in two key areas: (1) injury detection and post-injury behavior \citep{wall1979three, bolles1980perceptual, walters2023persistent, seymour2023post}, and (2) the storage and retrieval of skill memories in motor control. Neural ASMs offer a mechanism for injury (fault) detection by using a learned generative model that uses a hierarchy of prediction errors \citep{seymour2020hierarchical} to detect deviations and respond reactively. This complements other approaches, such as those using hand-designed generative models \citep{mahajan2025homeostasis}, that support more deliberate planning of post-injury behaviours e.g. injury investigation actions for information gain.

Conceptually, Neural ASMs imply the presence of a more central and hierarchically inferred representation of the injured state, which can guide appropriate motor responses. This contrasts with models that attribute post-injury behavior primarily to peripheral nociceptive hyperactivity \citep{walters2023persistent}. This central inference view is motivated by several considerations:
(1) Peripheral nociceptor signals are inherently noisy and unreliable indicators of injury, especially due to the variability in small-diameter unmyelinated C-fibre signaling \citep{debanne2004information};
(2) A single nociceptive channel cannot integrate other relevant sources of injury information, such as visual or physiological cues that do not directly modulate afferent nociceptive pathways \citep{hofle2010pain}; and
(3) Peripheral encoding alone lacks the temporal integration necessary to accumulate a history of observations and form a coherent, control-relevant representation of injury over time.

When it comes to motor control, on one hand, Neural ASMs appeals to the concepts such as the equilibrium point hypothesis \citep{feldman1995origin} and passive motion paradigm \citep{mohan2019muscleless}. Whilst, on the other hand is very compatible with \citet{schaal2007dynamics} where the dynamical systems policy sits atop the optimal control system, thus providing a possible unification of both approaches to motor control (Appendix 1, Fig. \ref{schaal_appendix}). 

The model's contextual inference of hidden states (Fig.~\ref{fig1}C, purple arrow) provides a potential mechanistic account of motor preparatory activity, hypothesized to initialise a dynamical system without explicitly encoding movement parameters \citep{churchland2010cortical, churchland2006neural, churchland2006preparatory, cisek2006preparing, fetz1992movement}. This mechanism parallels findings in RNNs, where initial sensitivity generates distinct sensorimotor sequences \citep{nishimoto2004learning, yamashita2008emergence}. However, traditional RNNs only have two timescales, a (faster) forward pass and a (slower) backward pass. Our model introduces an iterative inference process, also referred to as prospective configuration \citep{song2024inferring}, adding a third timescale to sensorimotor integration, which traditional RNNs lack. This iterative mechanism predicts a trade-off between speed and accuracy in memory recall and expression, consistent with experimental evidence of multiple timescales in the brain. Future work can explore comparisons with fast-weight RNNs \citep{ba2016using} which also explores the concept of adding a third timescale, but evolving weights at this third timescale, whereas our model evolves hidden states at inference timescale.


We further draw an analogy between Neural ASMs and System-1 thinking \citep{kahneman2011thinking}, described as a fast, intuitive, and largely unconscious mode of thinking but now here applied to fast and automatic motor behaviour. If the data is acquired by interacting with the environment under some policy instead of learning from demonstration or teleoperation then our approach is akin to modelling habits when viewed as a tendency to repeat previous actions \citep{thorndike2017animal, miller2019habits}, or a sequence of actions \citep{dezfouli2012habits, eltetHo2022tracking}, also referred to as procedural memory \citep{simor2019deconstructing}. A key distinction is that our model predicts both motor and associated sensory observations for the robot's stereotypical movements, similar to \citet{yamashita2008emergence, nishimoto2008learning}. This concept of "sensory forward model" is assumed to function in the inferior parietal cortex for the generation of skilled behaviours in humans and monkeys \citep{nishimoto2008learning}. 

Importantly, a speed-accuracy trade-off in habitual skill expression was observed in a human study by \citet{hardwick2019time}. Using a forced-response paradigm, participants were instructed to respond in sync with the final tone in a four-tone sequence following a stimulus. Varying the time between stimulus onset and the final tone, the authors found that for preparation times <300 ms, responses were at chance level due to insufficient stimulus processing. However, in habit-trained participants, habitual responses peaked between 300–600 ms, mirroring the speed-accuracy trade-off predicted by our model. Beyond 600 ms, responses were largely goal-directed. As our model lacks a goal-directed system, future work could explore the temporal dynamics of their competition. While their motor task is simpler than ours, the paradigm offers preliminary evidence supporting our model's predictions and a potential testbed in humans. Relatedly, rat studies by \citet{mizes2023dissociating, mizes2024role} involving multiple lever-press sequences found that motor cortex lesions impaired flexible, context-dependent skill expression but not single-sequence learning. Offline recall in our simulations is analogous to their working-memory-guided sequences, whereas online recall could correspond to sensory-guided ones.

\subsection*{Limitations}

We first discuss limitations specific to our simulation experiments and then those relevant to the Neural ASMs framework. Our experiments on fault detection and reactive correction are proof of concept. Fault detection using energy requires a threshold determined by the engineer based on data collected for specific skills/tasks. Tracking prediction errors helps isolate the effects of faults but not necessarily their causes, which remains an open research question. The reactive correction in our simulations employs a simple low-level position controller, which may not be energy-efficient or human-like. To simplify demonstration, we simulate a limited repertoire of two pick-and-place skills, although the framework supports more. We also find that the learning rate depends on the number of skills, consistent with observations in \citet{howard2015value} (see Appendix 3, Fig. \ref{howard_appendix}). For a detailed analysis of the memory and sequence capacity of tPC networks, please see \citet{tang2024sequential}.

Our contextual inference experiment is inspired by \citet{sheahan2016motor}, but it is not an exact replica of their human motor study. We omit the "no follow-through" condition, which lacks an explicit motor plan, based on several experimental findings \citep{gandolfo1996motor, howard2012gone, howard2013effect, howard2015value, sheahan2016motor} that no learning is observed under static cues. The inability to learn optimal trajectories to counteract motor perturbations from static cues is a nuance not captured in our simulation setup, where synthetic optimal trajectories are provided as demonstrations \citep{nishimoto2008learning, yamashita2008emergence}. This limitation is not specific to the models (e.g. traditional RNNs or tPC), but rather to the learning-from-demonstration paradigm. Nonetheless, it does not affect the finding that contextual inference of hidden states supports the separation and expression of learned skills.

Turning to our framework-level limitations:
First, concatenating sensory and motor observations as inputs/outputs, while aligned with prior works \citep{nishimoto2008learning, yamashita2008emergence}, may seem unconventional from an optimal feedback control standpoint \citep{todorov2002optimal}. Second, the goal of Neural ASMs is not to filter or estimate correct body state in the presence of sensor faults, unlike in \citet{lanillos2018adaptive}. Neural ASMs currently have fixed precisions; although they may act as filters during online recall, they cannot reliably infer hidden states from faulty sensors unless precision adapts dynamically. Third, temporal predictive coding does not estimate posterior uncertainty \citep{millidge2024}, which limits its use as a world model for planning. Fourth, tPC networks currently perform truncated BPTT over one time step \citep{tang2024learning}, restricting temporal resolution and sequence capacity. Lastly, ASMs and Neural ASMs are most useful under the assumption that skill-related movements are highly stereotyped \citep{pastor2012towards, pastor2013dynamic}, allowing reliable statistics to be collected from associated sensory traces. If sensory variability is too high, these statistics become unreliable.

\subsection*{Future Work}

Several promising directions emerge from our current work. While we employ offline recall in the current implementation, future work could explore online recall for the re-recognition of alternative skills in the repertoire. Improving reactive correction controllers for energy efficiency and human-likeness, such as using passive dynamic machines \citep{collins2005efficient} is another avenue. Expanding the skill repertoire beyond the simplified two-skill setup would better reflect real-world complexity and allow investigation of scaling dynamics, particularly learning rates as a function of skill number \citep{howard2015value}.

From a systems perspective, future work could explore implementing optimal feedback control as a low-level controller beneath Neural ASM policies \citep{schaal2007dynamics} (Appendix 1). One may also invert tPC networks (which approximate a Kalman filter) to implement Linear Quadratic Regulators (LQRs), bridging ASM with classical control architectures.

To enhance robustness under sensor faults, future work could enable dynamic adaptation of precisions or use alternative filtering strategies. Additionally, enabling Neural ASMs to track posterior variance e.g., via Monte Carlo Predictive Coding \citep{oliviers2024learning} would make them viable as planning-capable world models. For extending sequence capacity, our preliminary results suggest multi-timescale generative models are promising, similar to \citet{yamashita2008emergence}. Alternatively, rethinking backpropagation through time (BPTT) using eligibility traces e.g., e-prop \citep{bellec2020solution} or BP($\lambda$) \citep{pemberton2024bp} may offer more biologically plausible and efficient implementations for longer range credit assignment.

\subsection*{Conclusion}
In conclusion,  we advance the field of neurorobotics by proposing a novel human-like associative skill memory system for robots using biologically plausible local learning rules. The proposed system is unified in the sense that different operations minimise the same energy functional prescribed by the underlying embodied generative model. We have shown the utility of such a system in simple fault detection, reactive control and qualitative modelling of the role of contextual inference in skill memory expression in an arm robotics simulation test-bed, while making predictions about human motor preparation.

\section{Methods}

\subsection*{Model algorithm}
The sequential memory in Neural ASMs is modelled using a temporal Predictive Coding (tPC) network \citep{millidge2024, tang2024sequential}. Predictive coding models learn in a self-supervised fashion with the aim to best predict the incoming input based on its own learned generative model. The model evaluates the actual input against its prediction by determining the difference in activity in the respective error neurons and minimizes these `errors' through adjusting neural activities and synaptic weights, which corresponds to the processes of inference and learning, respectively \citep{Clark2015,bogacz2017tutorial}.\par

\begin{figure}[t]
\begin{center}
\includegraphics[width=\textwidth]{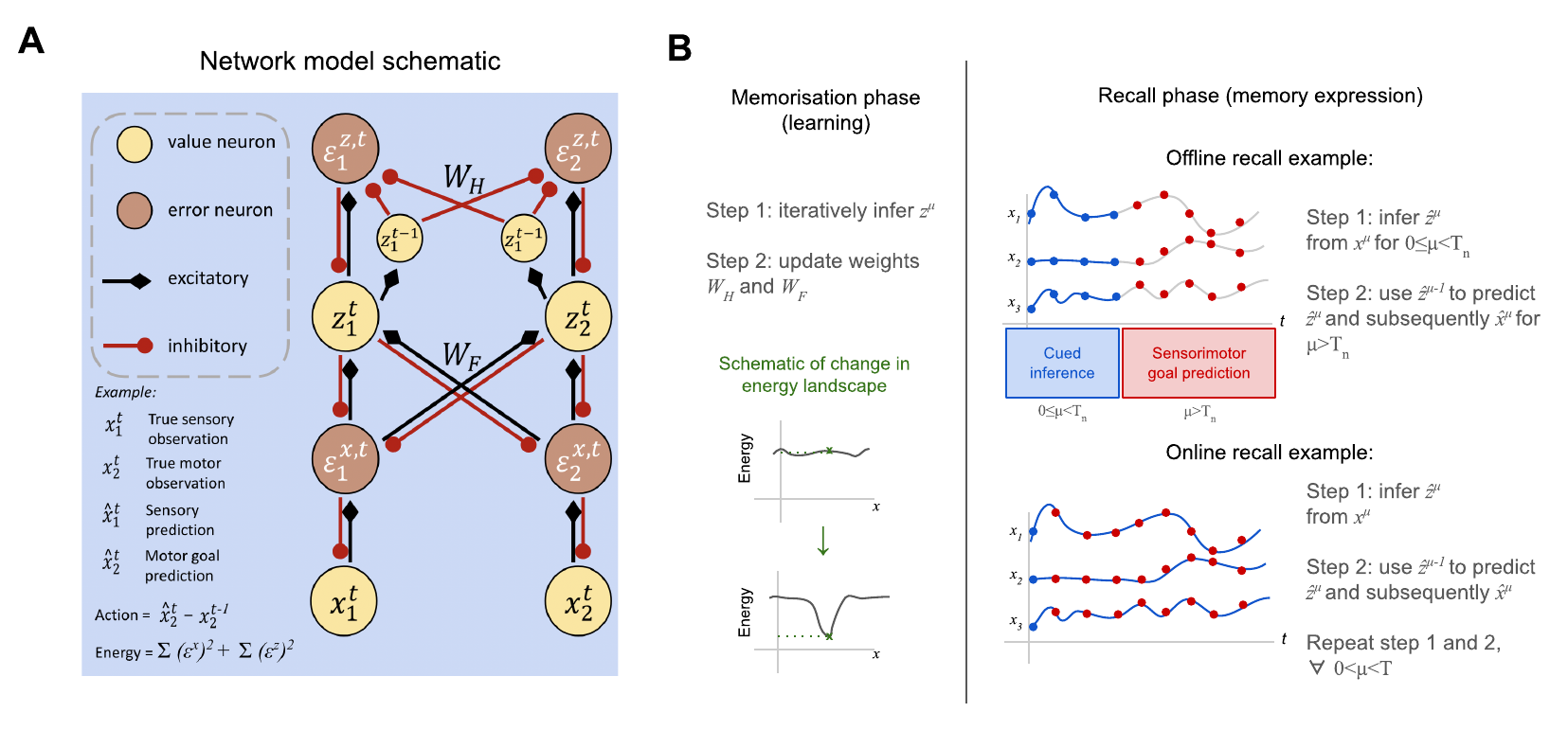}
\end{center}
\caption{(A) Neural network implementations of temporal predictive coding (tPC) model used in Results for simulations. The network is illustrated with a single sensory and motor observation inputs for didactic purposes (though in reality, the network will have multiple sensorimotor observations as inputs). (B) The model learns in the memorisation phase in a self-supervised manner and the weight updates change the energy landscape to store these memories as attractors in the energy landscape (which is crucial in recognising memorised skills). During the recall phase, the learned memories are expressed and can be either done offline (e.g. ballistic actions) or in an online manner where ground-truth observations are utilised to provide online-feedback through model inversion at each step.
} 
\label{fig_tpc_illustration}
\end{figure}

\begin{figure}[H]
\includegraphics[width=\textwidth]{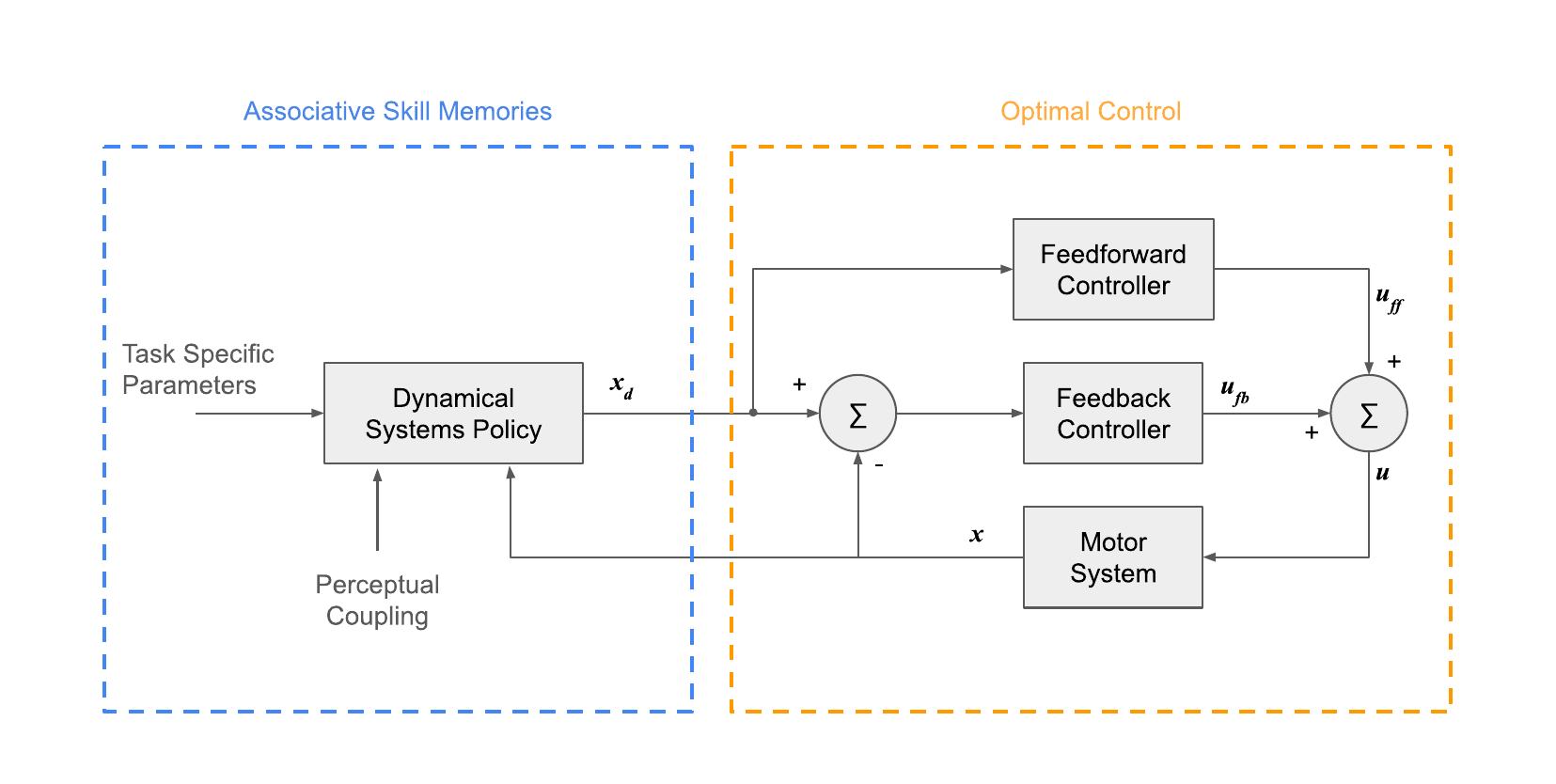}
\caption{Unifying view proposed by \citep{schaal2007dynamics}, where the dynamic systems policy from Associative Skill Memories can employ an optimal control-based low-level controller.} \label{schaal_appendix}
\end{figure}

In mathematical terms, the task of neural ASMs can be seen as learning a sequence of sensorimotor observations (such as motor coordinates and associated sensory events like haptic feedback etc) $(\mathbf{x}^t)_{t=0}^T$. This can be reduced to learning the dynamics in these sensorimotor observations for each skill, i.e. learning to associate each $\mathbf{x}^\mu$ with the next $\mathbf{x}^{\mu+1}$ $(\mu = 0, 1,..., T-1)$. We use a 2-layer tPC model, whose underlying graphical structure is that of a Hidden Markov Model (HMM). The lower layer of the tPC predicts the sensorimotor observations ($\mathbf{x}^{\mu}$) and the upper layer predicts the next hidden state ($\mathbf{z}^{\mu + 1}$). This predictive processing account loosely models the hierarchical processing of raw sensory inputs by the neocortex, where hidden
value neurons $\mathbf{z}^{\mu}$ models the brain’s internal neural responses to the sequential sensory inputs $\mathbf{x}^{\mu}$ (Fig. \ref{fig_tpc_illustration}A).

\par

The working of the tPC model in neural ASMs can be divided into two stages: (1) memorisation and (2) recall (Fig. \ref{fig_tpc_illustration}B). During memorisation, tPC tries to minimize the sum of squared errors at step $\mu$, with respect to the weights and the hidden activities:
\begin{equation}
F_{\mu}(\mathbf{z}^{\mu}, \mathbf{W}_H, \mathbf{W}_F) = \|\mathbf{z}^{\mu} - \mathbf{W}_H f(\mathbf{\hat{z}}^{\mu-1})\|_2^2 + \|\mathbf{x}^{\mu} - \mathbf{W}_F f(\mathbf{z}^{\mu})\|_2^2
\label{eqn:energy_f}
\end{equation}
where $\mathbf{W}_H$ governs the temporal prediction in the hidden state, $\mathbf{W}_F$ are the weights governing predictions from $\mathbf{z}^{\mu}$ to $\mathbf{x}^{\mu}$, with $\mathbf{\hat{z}}^{\mu-1}$ being the hidden state inferred at the previous time-step. During memorisation, the model first infers the hidden representation of the current sensorimotor observational input $\mathbf{x}^{\mu}$ by:
\begin{equation} 
\dot{\mathbf{z}}^{\mu} \propto -\frac{\partial F_{\mu}(\mathbf{z}^{\mu}, \mathbf{W}_H, \mathbf{W}_F)}{\partial \mathbf{z}^{\mu}} = -\varepsilon^{\mathbf{z}, \mu} + f'(\mathbf{z}^{\mu}) \odot \mathbf{W}_F^\top \varepsilon^{\mathbf{x}, \mu}
\label{eqn:iterative_inf}
\end{equation}
where $\odot$ denotes the element-wise product between two vectors, and $\varepsilon^{\mathbf{z}, \mu}$ and $\varepsilon^{\mathbf{x}, \mu}$ are defined as the hidden temporal prediction error $\mathbf{z}^{\mu} - \mathbf{W}_H f(\mathbf{\hat{z}}^{\mu-1})$ and the top-down error $\mathbf{x}^{\mu} - \mathbf{W}_F f(\mathbf{z}^{\mu})$, respectively. After $\mathbf{z}^{\mu}$ converges, $\mathbf{W}_H$ and $\mathbf{W}_F$ are updated following gradient descent on $F_{\mu}$:
\begin{equation}
\Delta \mathbf{W}_H \propto -\frac{\partial F_{\mu}(\mathbf{z}^{\mu}, \mathbf{W}_H, \mathbf{W}_F)}{\partial \mathbf{W}_H} = \varepsilon^{\mathbf{z}, \mu} f(\mathbf{\hat{z}}^{\mu-1})^\top
\label{eqn:learn_WH}
\end{equation}
\begin{equation}
\Delta \mathbf{W}_F \propto -\frac{\partial F_{\mu}(\mathbf{z}^{\mu}, \mathbf{W}_H, \mathbf{W}_F)}{\partial \mathbf{W}_F} = \varepsilon^{\mathbf{x}, \mu} f(\mathbf{z}^{\mu})^\top
\label{eqn:learn_WF}
\end{equation}
which are performed once for every presentation of the full sequence. Importantly, the converged $\mathbf{z}^{\mu}$ is then used as $\mathbf{\hat{z}}^{\mu}$ for the memorisation at time-step $\mu + 1$.

After memorisation (learning) is completed, the model enters the recall stage where all weights no longer change and the previously learned memories are expressed in response to certain cued input observations (also referred to as queries $q$). 
Note that during the recall phase, observation layer has no access to the correct patterns for the complete movement. Instead it needs to dynamically change its value to retrieve the memories to predict these sensorimotor observations. 
The sequential memories are recalled or expressed using the learned weights $\mathbf{W}_H$ and $\mathbf{W}_F$.  The loss thus becomes:
\begin{equation}
F_{\mu}(\mathbf{z}^{\mu}, \mathbf{\hat{x}}^{\mu}) = \|\mathbf{z}^{\mu} - \mathbf{W}_H f(\mathbf{\hat{z}}^{\mu-1})\|_2^2 + \|\mathbf{\hat{x}}^{\mu} - \mathbf{W}_F f(\mathbf{z}^{\mu})\|_2^2
\label{eqn:energy_f2}
\end{equation}
where $\mathbf{\hat{x}}^{\mu}$ denotes the activities of value neurons in the observation layer during recall. Both the hidden
and observation layer value neurons are updated to minimize the loss. The hidden neurons will follow similar
dynamics specified in Eq. \ref{eqn:iterative_inf}, whereas the
observation layer neurons are updated according to:
\begin{equation}
\dot{\mathbf{\hat{x}}}^{\mu} \propto -\frac{\partial F_{\mu}(\mathbf{z}^{\mu}, \mathbf{\hat{x}}^{\mu})}{\partial \mathbf{\hat{x}}^{\mu}} = -\varepsilon^{\mathbf{x}, \mu}
\label{eqn:iterative_pred}
\end{equation}
and the converged $\mathbf{\hat{x}}^{\mu}$ is the final retrieval.

In the case of sequential memory, there are two types of recall, offline and online. In case of offline recall, first $\mathbf{\hat{z}}^{\mu}$ is iteratively inferred from $T_n$ cued input ground-truth observations or queries $q = \mathbf{x}^{\mu}$, where $\mu = (0,1,..., T_n)$ (here, $0\leq T_n <T$). Once $\mathbf{\hat{z}}^{\mu}$ is converged, it is used to recall $\mathbf{\hat{x}}^{\mu}$ for $\mu > T_n$.  In case of online recall, we query the model with $q = \mathbf{x}^{\mu}$ (ground-truth), use the query to infer $\mathbf{\hat{z}}^{\mu}$, and then use $\mathbf{\hat{z}}^{\mu}$ for the recall the next time step $\mu+1$ for $\mu = 0, 1, ...,T-1$. This distinction is important in our results, here, we only present offline recall results for skill memory expression in ballistic movements.

\subsection*{Network training details}

The network details for the tPC network are as follows: The number of hidden units was 256 and the number of sensorimotor observation units depended on the task. The learning rate for the weight updates was $10^{-4}$ and the default learning iterations were $1000$ per skill, trained with a batch size of $1$. The iterative inference learning rate for hidden state update was $10^{-2}$ with default inference iterations set to $100$. The same hyperparameters were used for all simulation experiments. Kaiming uniform initialisation was used in hidden layers for all networks.

The S-to-M RNN and SM-to-SM RNN are sequence-to-sequence RNNs. They used the same number of hidden units, learning rate, learning iterations and batch size at the tPC network. They did not have iterative inference functionality, like tPC network. The input unit size is the number of sensory observations and output unit size is the number of motor observations. The skill memory expression experiment trains the RNNs to predict the output sequence using inputs only from the first time step to mimic the offline recall used in tPC. 

The sensorimotor sequences were Z-score normalised for each observation channel before providing as inputs to all neural networks. Outputs were again unnormalised to original units before movement.

\subsection*{Robot simulation details}

In our ASM simulations, Neural ASM operates at 2 Hz and the underlying low-level controller operates at 40 Hz and uses position control for simplicity. The dataset of demonstrations was generated in simulation by providing end-effector positions at each step, calculating joint angles with inverse kinematics and recording the sensory observations. This acts as a proxy for human demonstrations or teleoperation in real robots. The fault detection and reactive control experiment used 10 repetitions of two skills, with each sensorimotor sequence involving 25 observations (desired joint angles, end-effector positions, gripper positions, gripper open/close toggle, and sensed gripper forces and torques) over 15 time points (at 2 Hz). Each of the 10 repetitions had realistic noisy observations due to differences in inverse kinematics and sensed forces, etc. 

The systematic evaluation of fault detection and isolation experiment generates 980 faults by changing the joint that gets locked: joint \{1,2,3,4,5,6,7\}, the time of fault: t=\{1, 1.5, 2, 2.5, ..., 5.5\} and the degree of joint overshoot: \{-15$\degree$, -10$\degree$, -5$\degree$, 0$\degree$, 5$\degree$, 10$\degree$, 15$\degree$\}. The baseline method computes normalised errors for observation channel $i$ as follows: $\epsilon = (x_i^t - \bar{x_i}) / \sigma_i$. The sum of squared normalised errors is then used for fault detection, and the channel with maximum absolute normalised errors $|\epsilon|$ is used for fault isolation in our baseline.

The sensory and motor sequences used in the skill memory expression are presented in Appendix 2.

\section*{Author Contributions}
PM, IH and BS conceptualised the core idea, MT and TEL contributed with further discussions. PM designed and performed the simulation experiments. MT provided the initial code on the tPC network from his previous work. PM wrote the first draft of the manuscript with feedback from BS and IH. All authors contributed to editing and approving the manuscript.

\section*{Acknowledgements}
PM thanks Rafal Bogacz, Michael Browning, Gaspard Oliviers and anonymous reviewers from COSYNE 2024 and IWAI 2024 for their helpful feedback on earlier iterations of this work. PM and BS are funded by Wellcome Trust (214251/Z/18/Z, 203139/Z/16/Z and 203139/A/16/Z), IITP (MSIT 2019-0-01371) and JSPS (22H04998). MT is supported by E.P. Abraham
Scholarship in the Chemical, Biological/Life and Medical
Sciences. TEL was supported by the Gruber Science Fellowship and the Interdepartmental Neuroscience Program at Yale university, which is funded by T32 NS041228 from the National Institute of Neurological Disorders and Stroke. This research was also partly supported by the NIHR Oxford Health Biomedical Research Centre (NIHR203316). The views expressed are those of the author(s) and not necessarily those of the NIHR or the Department of Health and Social Care.

The authors have no conflicts of interest to declare.

\bibliography{references}

\section*{Appendix 1: Unifying view on dynamical systems vs optimal control}

\section*{Appendix 2: Demonstration data used in skill memory expression task}

\begin{figure}[H]
\includegraphics[width=\textwidth]{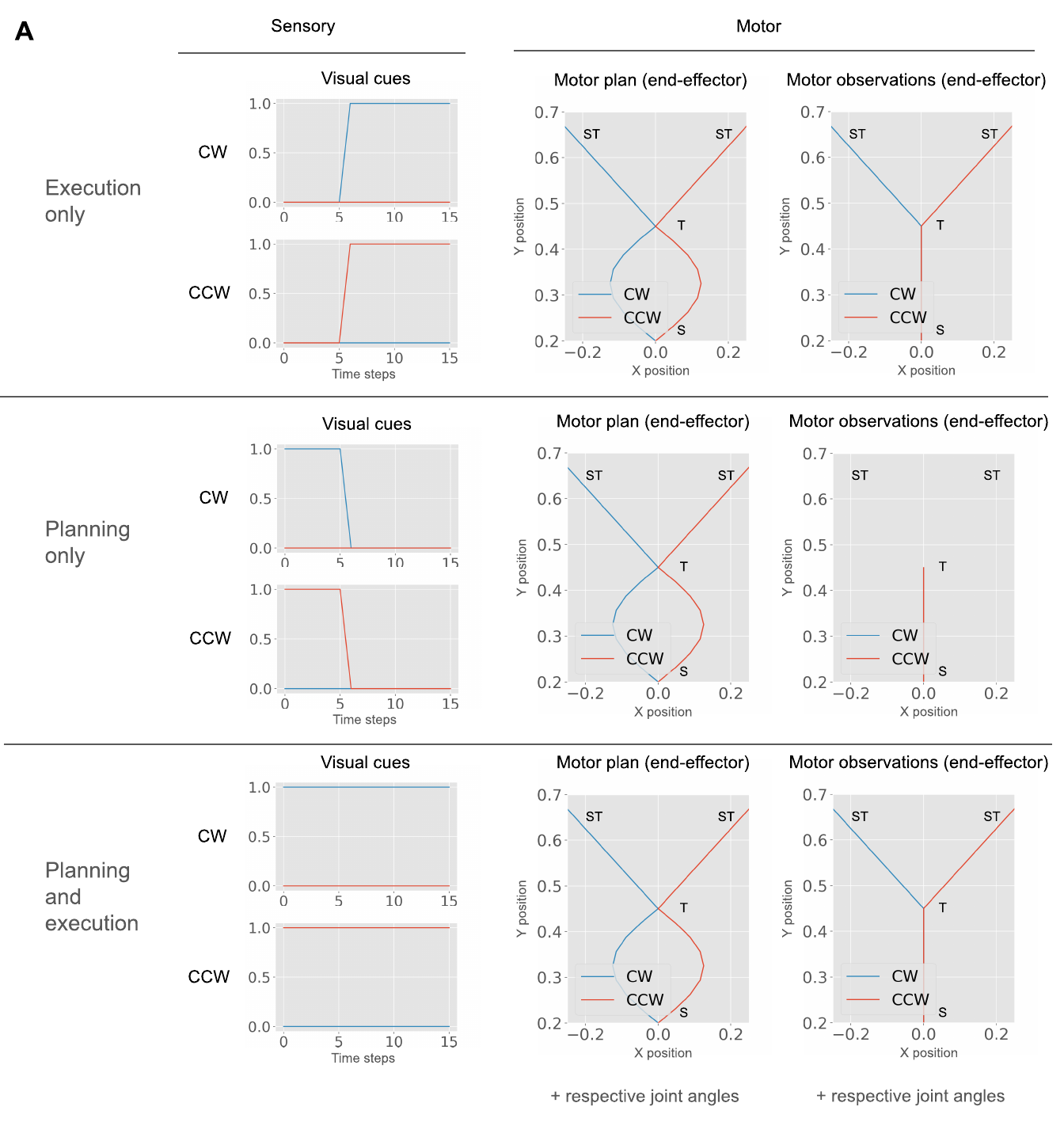}
\caption{Plots of the sensory and (hypothetically optimal) motor sequences used for demonstrations in robot experiments on skill memory expression inspired by \citet{sheahan2016motor}.} \label{sheahan_appendix}
\end{figure}

\section*{Appendix 3: Learning rate varies with number of skills}

\begin{figure}[H]
\includegraphics[width=\textwidth]{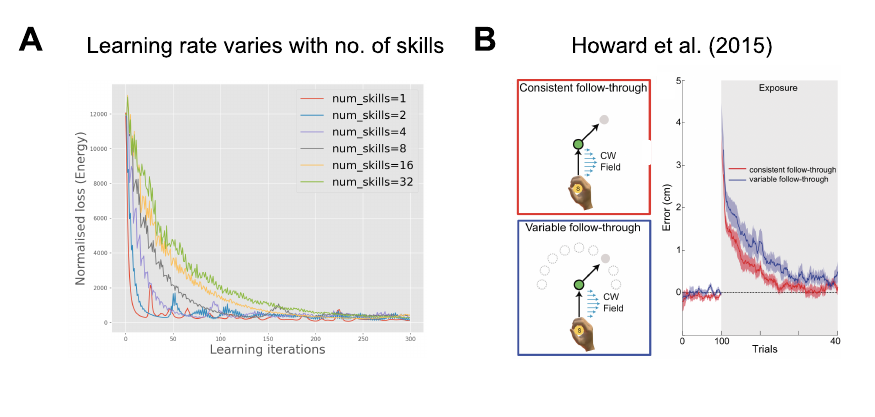}
\caption{Lesser the number of distinct skills to memorise, faster is the learning rate of our model, as seen in energy (normalised loss) over epochs. This potentially explains why consistent follow-throughs improve learning rates as variable follow-throughs can split the learning into different skill memories rather than a single memory, as observed by \citet{howard2015value}.} \label{howard_appendix}
\end{figure}

\end{document}